# Hybrid Distillation with CoT Guidance for Edge-Drone Control Code Generation

Yizhan Feng[1,4], Hichem Snoussi[1,4], Yuhang Wang[2], Jing Teng[2], Abel Cherouat[3,4] and Tian Wang[5]

[1] UR-LIST3N, University of Technology of Troyes, Troyes, France
E-mail: hichem.snoussi@utt.fr
[2] Institute of Artificial Intelligence, North China Electric Power University, Beijing, China
[3] UR-GAMMA3, University of Technology of Troyes, Troyes, France
[4] *EUt+, Data Science Lab, European Union*
[5] Institute of Artificial Intelligence, Beihang University, Beijing, China

**Summary:** With large language models demonstrating significant potential in code generation tasks, their application to onboard control of resource-constrained Unmanned Aerial Vehicles has emerged as an important research direction. However, a notable contradiction exists between the high resource consumption of large models and the real-time, lightweight requirements of UAV platforms. This paper proposes an integrated approach that combines knowledge distillation, chain-of-thought guidance, and supervised fine-tuning for UAV multi-SDK control tasks, aiming to efficiently transfer complex reasoning and code generation capabilities to smaller models. Firstly, a high-quality dataset covering various mainstream UAV SDKs is constructed, featuring instruction-code-reasoning chains, and incorporates counterfactual negative samples for data augmentation, guiding the model to learn the end-to-end logic from instruction parsing to code generation. Secondly, leveraging DeepSeek-Coder-V2-Lite quantized via QLoRA as the teacher model, and based on a hybrid black-box and white-box distillation strategy, high-quality chain-of-thought soft labels are generated. These are combined with a weighted cross-entropy loss using hard labels to transfer complex reasoning capabilities to the smaller student model. Finally, through prompt tuning engineering optimized for the UAV control scenario, the model's performance on core tasks such as SDK type recognition and function call matching is enhanced. Experimental results indicate that the distilled lightweight model (parameters≤1B) maintains high code generation accuracy while achieving significant improvements in deployment and inference efficiency, effectively demonstrating the feasibility and superiority of our approach in achieving precise and lightweight intelligent control for UAVs.

## 1. Introduction

Large language models have achieved breakthrough progress in the field of natural language processing, and their powerful language understanding and generation capabilities are rapidly penetrating critical areas such as robotics and autonomous driving [1, 2]. For instance, the ChatGPT has attracted widespread attention due to their strong general-purpose capabilities [3]; the DeepSeek series has become an important open-source choice owing to its excellent overall performance and advantages in long-context handling [4]; the Llama series demonstrates significant potential in specific domain applications thanks to its open ecosystem and extensive community fine-tuned versions [5]; and the Qwen series offers a full-stack model family ranging from large-scale to lightweight versions, excelling particularly in code understanding and tool calling [6]. These models collectively push the boundaries of technology, providing a rich foundation of models for downstream applications. In the context of UAV applications, researchers have begun exploring the use of LLMs for high-level task planning, natural language interaction, and the automatic generation of control code, opening up new possibilities for creating more intelligent and user-friendly autonomous UAV systems [7].

However, directly deploying the most advanced large models on resource-constrained UAV edge devices poses severe challenges. These models typically contain tens or hundreds of billions of parameters, demanding extremely high computational power, memory, and energy consumption. In contrast, UAV platforms are constrained by size, weight, and power limitations, resulting in limited onboard computing capacity. Consequently, most current research remains within a "ground-side" paradigm, where powerful LLMs running on cloud servers or ground stations perform asynchronous task planning and analysis, sending resultant commands to the UAV [8]. This mode is inevitably hampered by network latency, bandwidth instability, and data privacy risks, making it difficult to meet the requirements for low-latency, high-reliability control tasks such as real-time obstacle avoidance and dynamic path planning for UAVs.

To break through this bottleneck and enable real-time intelligent control with LLMs on UAV onboard platforms, model lightweighting techniques have become crucial. Knowledge Distillation (KD) has emerged as an effective model compression method [9, 10]. Its core idea follows a "teacher-student" framework, transferring the knowledge encapsulated within a large, complex pre-trained "teacher model" to a lightweight "student model". The classical KD process often introduces a temperature parameter into the Softmax function to generate smoothed "soft labels". These soft labels contain not only the teacher model's confidence in the correct category but also reveal richer knowledge, such as similarity relationships between different categories, offering far more instructive value than original "hard labels". The training objective for the student model is a weighted combination of two loss functions: one is the distillation loss, consistent with traditional supervised learning but based on the teacher's soft labels, typically measured using Kullback-Leibler divergence to quantify the distribution difference between the student's output and the teacher's soft labels; the other is the student loss, based on the ground-truth labels [10]. Through this joint optimization, the student model can learn not only specific task knowledge but also mimic the generalization "thinking" of the teacher model, thereby maintaining performance close to the teacher's even with a significant reduction in parameters [11, 12].

The Chain-of-Thought (CoT) technique is key to enhancing the multi-step reasoning abilities of large models [13]. The core of CoT lies in simulating human thinking processes. Through specific prompt engineering techniques, it guides the model to explicitly generate a sequence of intermediate reasoning steps before producing the final answer. For example, when faced with a composite mathematical word problem, CoT guides the model to first decompose the problem, gradually compute intermediate results, and finally arrive at the answer, rather than performing direct end-to-end output. This approach breaks down complex reasoning tasks into more manageable sub-problems, significantly improving the model's accuracy not only in tasks like arithmetic, commonsense reasoning, and symbolic manipulation but also greatly enhancing the interpretability and controllability of the model's decision-making process, as errors can be traced to specific steps [14]. The methodology has evolved to include the Program of Thoughts—an approach that translates natural language reasoning steps into executable code to enhance computational accuracy, thus enabling adaptation to more complex reasoning scenarios.

Fortunately, several mainstream LLMs (e.g., DeepSeek, Llama, Qwen) have been open-sourced, and lightweight versions specifically optimized for code generation tasks have been released, providing a preliminary model foundation for deployment on resource-constrained devices. Our preliminary research work [15, 16] has already validated the feasibility of deploying these code-optimized lightweight LLMs for multi-task UAV control in simulated environments. However, discrepancies exist between simulation environments and the real physical world. Furthermore, to adapt to the wide variety of UAV SDKs with their respective unique features, and to meet the stringent requirements of small platforms with extremely limited computational resources, deeper compression and targeted fine-tuning of these existing LLMs are still necessary. Therefore, this paper proposes a collaborative optimization framework integrating knowledge distillation and chain-of-thought reasoning. It aims to further compress the model while precisely transferring the complex multi-step reasoning capabilities from the teacher model to the student model, ultimately realizing an lightweight onboard model suitable for various UAV platforms, capable of real-time code generation and interactive control.

## 2. Methodology

We propose a knowledge distillation framework for UAV control tasks, comprising three core stages: 1. construction of a high-quality chain-of-thought instruction dataset based on multiple mainstream UAV SDKs; 2. a hybrid distillation strategy to transfer complex reasoning capabilities from a teacher model to a lightweight student model; and 3. the fine-tuning integrated with prompt engineering to further enhance instruction adherence and code generation performance. The overall framework of the proposed method is illustrated in **Fig. 1**.

In the data construction phase for UAV control, we build a large-scale, high-quality multi-SDK instruction dataset. Data sources cover official Python API documentation from eight mainstream UAV SDKs: TelloPy, olympe, MAVSDK, AirSDK, Skydio, dronekit, mavproxy, and Wingtra, ensuring the model's adaptability to diverse hardware platforms and software ecosystems. Dataset construction adopts a multi-level annotation strategy: automatically generating basic instruction-code pairs from API documentation, enhancing them with detailed chain-of-thought reasoning processes through the teacher model, and finally conducting manual verification to ensure data quality. Each training sample strictly follows the <instruct-think-code> triplet structure, where natural language instructions simulate real operational scenarios requiring composite tasks involving 3-5 API calls; the thought chain elaborately demonstrates the complete reasoning process from instruction parsing, SDK selection, function call sequence determination to parameter validation; the final code implementation strictly adheres to each SDK's syntax specifications and incorporates comprehensive exception handling mechanisms. This structured data ensures the model learns the end-to-end closed-loop logic from semantic understanding to code generation. Furthermore, inspired by counterfactual distillation research [17], we employ multi-perspective CoT technology to enhance model robustness. Specifically, through prompt engineering, we guide the teacher model to generate both positive CoT and negative CoT (counterarguments for incorrect options), enabling the model to understand task logic from both positive and negative angles, thereby enhancing its comprehension of causal relationships.

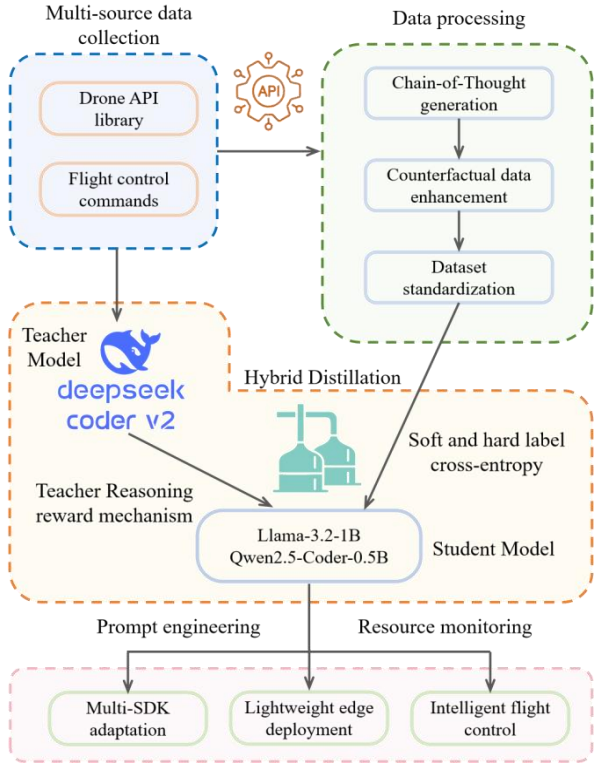

**Fig. 1.** Overview of the proposed methodological framework.

In the model selection and distillation strategy phase, to meet the stringent lightweight requirements of UAV onboard platforms, this work selects two types of high-performance, low-parameter open-source models as student bases: Llama-3.2-1B-Instruct and Qwen2.5-Coder-0.5B-Instruct [5, 6]. The teacher model employs DeepSeek-Coder-V2-Lite fine-tuned with QLoRA quantization. QLoRA is an efficient fine-tuning technique that significantly reduces memory footprint while maintaining model performance through low-rank adaptation and gradient backpropagation optimization, making it particularly suitable for resource-constrained edge computing scenarios [18]. DeepSeek-Coder-V2 is a large language model specifically designed for code generation tasks, employing a mixture-of-experts architecture and code-specific training corpus. Its lite version substantially reduces computational requirements while maintaining powerful code capabilities [19]. We combine the flexibility of black-box distillation with the precision of white-box distillation to maximize knowledge transfer efficiency. Black-box distillation utilizes the teacher model to generate high-quality CoT reasoning processes as soft labels, employing Softmax functions to produce smoothed probability distributions, which enables the student model to learn the teacher's decision-making process. Meanwhile, white-box distillation aligns the intermediate representations of teacher and student models by minimizing the mean squared error between hidden states, allowing the student model to learn richer semantic representations.

In the fine-tuning optimization phase, this study incorporates prompt engineering methods to enhance fine-tuning effectiveness. Prompt engineering significantly influences model output quality and generalization capability through careful design of input templates and instruction formats. We design structured prompt templates that explicitly specify SDK types, API functions, and parameter constraints. The fine-tuning process focuses on three core tasks: SDK type identification, function call matching, and parameter integrity verification. Throughout the training process, we continuously monitor key metrics including loading time, inference speed, and resource utilization, ensuring the model meets real-time control requirements.

## 3. Experiments

To comprehensively evaluate the effectiveness of the proposed large model distillation method for UAV control tasks, the experiments were conducted on a desktop environment equipped with an NVIDIA RTX 4090 GPU. Training stability is the foundation for ensuring reproducible model performance. Throughout the training process, the final training loss of the Llama-3.2-1B model stabilized below 0.5, while the Qwen2.5-Coder model's loss converged below 0.8, both showing a stable and consistently decreasing trend without significant fluctuations. At the token level, the models achieved an accuracy of over 80% on the entire training set. These results demonstrate that the hybrid distillation and counterfactual data augmentation strategy effectively guides the small models in learning knowledge from the teacher model, leading to a robust training process with good convergence.

The results for efficient inference optimization are significant. As shown in **Table 1**, the distilled small models ($\leq 1$ Bparameters) achieved an order-of-magnitude inference acceleration, with a generation speed reaching 400-595 tokens/s, far exceeding the speed of the teacher model (9.85 tokens/s) . In terms of loading time, the optimized models were significantly shortened relative to their baseline versions. The loading time for the Qwen_0.5B_Counter was as low as 1.71s, fully meeting the real-time requirement for rapid startup in UAV systems. Regarding memory usage, the peak GPU memory footprint of all distilled models was controlled within 5.1GB, and the system runtime memory increase remained stable at a low level of approximately 1.4GB. This metric fully demonstrates that the distilled models satisfy the stringent resource constraints of UAV onboard platforms.

The incorporation of counterfactual samples has also endowed the models with the ability to identify infeasible instructions. In specialized tests, for instructions beyond their capability scope, the models could accurately recognize them and return appropriate prompt messages (e.g., Current SDK does not support this function), rather than generating erroneous code or invalid operations. This error-handling mechanism significantly enhances the reliability and safety of the model in practical applications, avoiding potential flight risks caused by erroneous instructions.

Table 1. Model resource efficiency performance.

| Model | Loading Time (s) | Generation Speed (tokens/s) | GPU Memory (GB) | Runtime Increase (GB) |
|---|---|---|---|---|
| QLoRA-DeepSeek | 19.15 | 9.85 | 10.54 | 2.20 |
| Llama-1B (Baseline) | 12.74 | 586.82 | 4.43 | 1.45 |
| Qwen-0.5B (Baseline) | 6.17 | 400.91 | 5.03 | 1.40 |
| Llama-1B-Counter | 2.39 | 595.6 | 4.3 | 1.43 |
| Qwen-0.5B-Counter | 1.71 | 400.93 | 5.08 | 1.42 |

## 4. Conclusion

We propose a LLM distillation method for UAV control tasks. By integrating hybrid distillation, chain-of-thought guidance, and counterfactual data augmentation, it has successfully transferred the complex reasoning and code generation capabilities of large models to lightweight models with fewer than 1B parameters. Experimental results indicate that the distilled models achieve an effective balance between performance and efficiency, the training process is stable, functional coverage is comprehensive, and the models support end-to-end tasks across multiple SDKs. More importantly, the resource efficiency of the models is significantly improved, with markedly reduced loading times, inference speeds meeting real-time control requirements, and minimal GPU memory and RAM usage, fully satisfying the stringent constraints of UAV onboard platforms. This study provides a feasible technical solution for deploying large models on resource-limited UAV edge devices, enabling real-time and reliable human-machine interactive intelligent control in complex scenarios.

## Acknowledgements

This work has been partially funded by BPI DreamScanner project.